# MSMix: An Interpolation-Based Text Data Augmentation Method Manifold Swap Mixup

Mao Ye[1,2]  Haitao Wang[1]  Zheqian Chen[1]

[1]AI laboratory, Yiwise, Hangzhou, China
yemao@yiwise.com  wanghaitao@yiwise.com  chenzheqian@yiwise.com
[2]College of Computer Science and Technology, Zhejiang University, Hangzhou, China

## *Abstract*

*To solve the problem of poor performance of deep neural network models due to insufficient data, a simple yet effective interpolation-based data augmentation method is proposed: MSMix (Manifold Swap Mixup). This method feeds two different samples to the same deep neural network model, and then randomly select a specific layer and partially replace hidden features at that layer of one of the samples by the counterpart of the other. The mixed hidden features are fed to the model and go through the rest of the network. Two different selection strategies are also proposed to obtain richer hidden representation. Experiments are conducted on three Chinese intention recognition datasets, and the results show that the MSMix method achieves better results than other methods in both full-sample and small-sample configurations.*

## 1. Introduction

In recent years, deep neural networks have made breakthroughs in supervised configurations for most natural language processing tasks. In particular, the birth of the large-scale pre-training model BERT[1] has enabled the pre-training plus fine-tuning paradigm to prevail. Although large-scale pre-training models retain more a priori knowledge, they still require a large amount of labeled data for fine-tuning to achieve better results in a specific task or domain. In real-world scenarios, it is more common to have a large amount of unlabeled data and a small amount of labeled data. A simple and straightforward way to improve the performance of deep neural networks is to manually label these unlabeled data, which requires a lot of human, financial, and time resources, and for data in specific specialized fields, even requires specialized expertise to label. With a small amount of labeled data, deep neural network models can easily overfit the available labeled data, capturing only less information that can help in downstream tasks, resulting in poor model performance.

In order to address the issue of poor model performance with limited sample data, a series of methods have been summarized and collected[2]. Among them, the classic solution at the data level is data augmentation, which generates additional data for model training based on prior knowledge. Generally, existing text data augmentation methods can be classified into two categories: input-level augmentation and hidden representation-level augmentation. Input-level augmentation modifies the original text at the character, word, or sentence level using specific strategies to generate new, similar texts while preserving the original labels, as in EDA[3]. The augmented data generated by these methods are perceivable and understandable by humans. Hidden representation-level augmentation operates on the intermediate hidden representations of the text within the deep neural model, as in TMix[4]. This method utilizes interpolation-based techniques to obtain new intermediate representations, which are then used as inputs for subsequent layers. The augmented data generated by these methods are virtual samples that are difficult for humans to intuitively understand.

Inspired by SwapMix[5] and Manifold Mixup[6], we propose a novel simple and effective data enhancement method: MSMix (Manifold Swap Mixup). The MSMix method is a Mixup method that acts on the space of textual hidden representations, and the hidden representations of two samples input to the depth model. The MSMix method is a Mixup method that acts on the hidden representation space of text hidden representations, where the hidden representations of two samples input to the depth model are manipulated to replace some of their dimensions. Three different dimensional replacement strategies are also proposed to explore more effective replacement methods. Extensive experiments are conducted on three Chinese intention recognition datasets to confirm the effectiveness of the MSMix method proposed in this paper.

## 2. RELATED WORK

### 2.1. Traditional Text Data Augmentation

Many of the existing methods for text data augmentation are inspired by related methods in the field of computer vision[7]. The EDA[3] utilizes four operations to generate new augmented data: synonym replacement, random insertion, random swapping of order, and deletion. However, the efficacy of EDA-generated data is unstable, as this method randomly changes some words in a sentence, which may have a significant impact on semantics or even introduce errors in grammar. Nevertheless, EDA still preserves the original labels, leading to strong misguidance for the model. A technique called "back-translation"[8] translates the original text into an intermediate language (e.g., Chinese sentences translated into English) and then back into the original language to obtain augmented data. SSMBA[9] use masked language models to generate augmented data. This approach first masks some tokens in the original text at random, and then uses a masked language model to predict the masked tokens, or directly generates samples using large-scale language models[10]. Back-translation and the use of language models rely heavily on prior knowledge, and the generated augmented samples tend to converge, approaching the pre-training data of the language model. In addition, some research works[11, 12] use certain strategies to select and replace or delete unimportant words for subsequent tasks to achieve the goal of data augmentation.

### 2.2. Interpolation-based Data Augmentation

The Mixup[13] is an interpolation-based data augmentation method. Due to its superior performance, it first gained popularity in the computer vision field and was later applied to natural language processing, spawning many variants. Data augmentation based on the hidden space is mostly based on Mixup.

To improve the generalization ability of the model, Mixup linearly blends two different samples and applies the same operation to the labels. Specifically, for a given labeled sample dataset $D(X,Y)$, the following operations are performed:

$$x' = \lambda x_i + (1-\lambda) x_j \tag{1}$$

$$y' = \lambda y_i + (1-\lambda) y_j \qquad (2)$$

where $(x_i, y_i)$ and $(x_j, y_j)$, $x_i, x_j \in X$, $y_i, y_j \in Y$ denote two samples randomly selected from the set of sample data with labels D, and $\lambda \in [0,1]$ is taken from the beta distribution.

Mixup is an operation at the pixel level of the original image, which yields an unnatural image, based on which CutMix[14] crops and stitches two images at the image block level. Recently, SwapMix[5] trains the model by replacing features of irrelevant visual context objects with features of other objects in the dataset to generate new samples, thus reducing the effect of irrelevant visual context on the final prediction.

In the field of NLP, there are also studies on performing Mixup operations on the original text. For example, SSMix[15] replaces unimportant words in the current sentence with important words in another text based on the importance of word vectors in the sentence.

Manifold Mixup[6] applies the Mixup technique to the hidden representations of a model. They believe that the hidden representations contain higher-order semantic information, so linear interpolation is performed on their dimensions, resulting in more meaningful virtual samples. As a result, many excellent research results have emerged in the NLP field, such as TMix[4], which, combined with other data augmentation techniques and self-supervised learning frameworks, developed the MixText model. In Mixup-Transformer[16], Mixup was applied to the output of the last layer of a Transformer-based model. DMix[17] believes that fusing two samples extracted using a certain strategy will yield better results than random sampling. Therefore, they construct a set of samples whose hyperbolic distance from a sample in the dataset exceeds a set threshold, and randomly extract a sample from the set for Mixup operation. The study also sets λ as a learnable parameter matrix. DoubleMix[18] first extracts a sample, performs traditional data augmentation on it, then performs Mixup operation on the two samples, and finally performs Mixup operation on the Mixup result and the original sample.

### 2.3. Regularization

The study[19] argues that Mixup can be equated to a form of regularization and derives a model of how it can be regularized. The MSMix proposed in this paper is based on the assumption that each dimension of the hidden layer represents a virtual feature and that each downstream classification task results in a corresponding significant feature dimension. If some dimensions are replaced randomly or strategically, it can be equated to apply noise to the hidden layer, and the noise is not randomly generated based on some distribution, but is extracted from the hidden layer of another real sample, which acts as a regularization and improves the robustness of the model.

The MSMix proposed in this paper can also be regarded as a variant of Dropout[20], where Dropout is a random dropout of neurons, and the study[21] is a random dropout of a dimension, while MSMix is a replacement of dimensions.

### 3. PROPOSED WORK

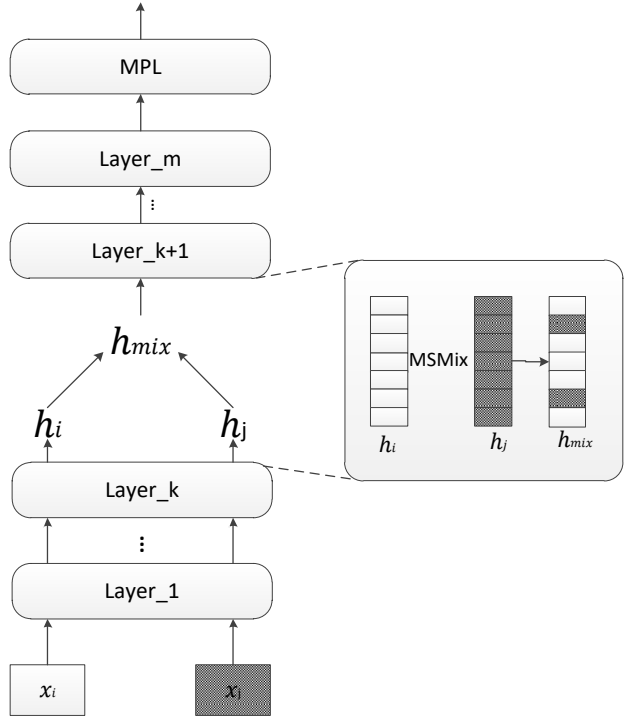

Figure 1 Structure of MSMix

### 3.1. Model Overview

The MSMix model architecture proposed in this article is shown in Figure 1. The training dataset is represented as $D=(X,Y)$, where $x_i, x_j \in X$ represent different textual sentences sampled from the training set, and $y_i, y_j \in Y$ represent the labels corresponding to the sampled samples, with $i, j \in [0, N]$, where N is the total number of samples and $i \neq j$. $k \in [0, m]$ represents a certain layer in the deep neural network, where $k = 0$ represents the input layer. $m$ is the maximum number of layers in the deep neural network, such as the 12 layers in the Bert model.

Before the training samples enter the deep neural network model, the output $h^k$ of the kth layer in the deep neural network is randomly selected as the input of MSMix. The two text segments $x_i, x_j$ are input into the deep neural network model and pass through k layers to obtain $h_i^k, h_j^k$. This process can be expressed by the following mathematical formula:

$$h_i^k = f_{\theta_k}\left(h_i^{k-1}\right), k \in [1, m] \qquad (3)$$

$$h_j^k = f_{\theta_k}\left(h_j^{k-1}\right), k \in [1, m] \qquad (4)$$

where $f(\cdot)$ denotes the deep neural network and $\theta_k$ denotes the parameters of the k-th layer.

After obtaining $h_i^k, h_j^k$, MSMix them (to be specified in 3.2) to obtain $h_{mix}^k$, the process is shown in the following equation:

$$h_{mix}^k = MSMix_\lambda \left( h_i^k, h_j^k \right) \tag{5}$$

then $h_{mix}^k$ is input to the subsequent hidden layer, and finally the model prediction result $\hat{y}$ is obtained after the fully connected layer.

For the labels of the synthetic samples, this paper uses the original Mixup operation to fuse the labels:

$$y' = \lambda y_i + (1-\lambda) y_j \tag{6}$$

$$\lambda \sim Beta(\alpha, \alpha) \tag{7}$$

In this paper, we use cross-entropy loss to optimize the model parameters:

$$\mathcal{L}(y', \hat{y}) = -\sum_{i=1}^{n} y_i' \log y_i \tag{8}$$

### 3.2. MSMix

SwapMix[5] applies the technique of replacing features of unrelated visual contextual objects with features of other objects in the dataset to generate augmented data. Since SwapMix can extract features of unrelated visual contextual objects from image data that has a large amount of labeled information, while text data does not have such rich labeling information, this paper only applies similar methods at the hidden representation level to select and replace dimensions of hidden representations, and explores strategies for selecting and replacing dimensions. The idea is as follows: determine $\lambda$ by Beta distribution, randomly select $p$ dimensions in the hidden representation $h_i^k$, and replace them with the corresponding $p$ dimensions in $h_j^k$. And $p = \lfloor \lambda * d \rfloor$, where $d$ denote the feature dimensions of the hidden layer output. The method is denoted as MSMix-base. the specific expression for obtaining $h_{mix}^k$ is shown below:

$$h_{mix}^k = M \odot h_i^k + (1-M) \odot h_j^k \tag{9}$$

where $M \in \{0,1\}^{I \times d}$ represents the mask matrix for selecting dimensions of the hidden representation, and $I$ is the maximum length limit of the input tokens; specifically, $M$ has $p \times I$ zeros and the rest are ones. $\odot$ represents element-wise multiplication.

Building on this, we further hypothesize that adding some restrictions or selection strategies to the dimension selection of hidden representations will lead to better performance. Therefore, in this paper, we propose two dimension selection strategies: MSMix-A and MSMix-B.

MSMix-A: The values of each dimension in $h_i^k, h_j^k$ are multiplied correspondingly and then the absolute values are taken, and then the values of the $d$ dimensions are arranged in descending order, and the top $p$ dimensions are selected as the replacement dimensions. The specific process is as follows:

$$C = \left| h_i^k \odot h_j^k \right| \tag{10}$$

$$M = argmax_p(C) \tag{11}$$

where $C$ denotes the result obtained after multiplying the two hidden representations corresponding to the absolute value, $argmax_p(\cdot)$ denotes the dimension of the largest $p$ returned, and the corresponding dimension in the mask matrix M is set to 0. After obtaining the mask matrix M, we use Equation (9) to calculate $h_{mix}^k$ and input it to the subsequent network layer for calculation.

MSMix-B: Taking the absolute values of each dimension in $h_i^k$, then selecting the smallest $q$ dimensions, and sorting the corresponding $q$ dimensions in $h_j^k$. Next, selecting the largest $p$ dimensions from $h_j^k$ and replacing them with the corresponding dimensions in $h_i^k$. The specific process is as follows:

$$M_q = argmin_q\left(\left|h_i^k\right|\right) \quad (12)$$

$$M = argmax_p\left(\overline{M_q} \odot h_j^k\right) \quad (13)$$

where $argmin_q(\cdot)$ denotes the $q$ dimensions with the smallest value selected, $M_q$ is the mask matrix obtained by this process; $\overline{M_q}$ denotes the result of 0, 1 interchange in $M_q$, i.e., $q$ dimensions are retained; $q \geq p$ in the above equation. After equations (12) and (13) to obtain the mask matrix M using equation (9) to calculate $h_{mix}^k$, which is input to the subsequent network layer for calculation.

## 4. EXPERIMENTS

### 4.1. Datasets

This paper conducts experiments on three Chinese datasets: YiwiseIC, SMP2017-ECDT[22] and CrossWOZ-IC. YiwiseIC is an intent classification Chinese dataset collected by Hangzhou Yiwise Intelligence Technology Co., Ltd. in combination with business scenarios. SMP2017-ECDT is a dataset used in the 6th National Conference on Social Media Processing for Chinese human-machine dialogue technology evaluation, and this paper uses the intent classification part of the dataset. CrossWOZ[23] is a large-scale cross-domain dialogue system dataset with rich data and complex structure. Since intent classification is inevitably involved in human-machine dialogue systems, when conducting experiments using the CrossWOZ dataset, this paper first processed the data, extracted the text spoken by humans and the corresponding intent labels for experimentation, and referred to the extracted dataset as CrossWOZ-IC.

In order to verify the effectiveness of the MSMix method proposed in this paper in the small-sample scenario, the above three Chinese datasets were sampled in the experiment (details can be found in Section 4.3.2), and three small-sample datasets were obtained, which are respectively referred to as YiwiseIC_FS, SMP2017-ECDT_FS, and CrossWOZ-IC_FS. The specific information of each dataset is shown in Table 1.

Table 1 Dataset detail information

| dataset | Number of samples | Number of classes |
|---|---|---|
| YiwiseIC_train | 28509 | 134 |
| YiwiseIC_val | 1643 | 134 |
| YiwiseIC_test | 2705 | 134 |
| SMP2017-ECDT_train | 2299 | 31 |
| SMP2017-ECDT_val | 716 | 31 |
| SMP2017-ECDT_test | 667 | 31 |
| CrossWOZ-IC_train | 42346 | 9 |
| CrossWOZ-IC_val | 4229 | 9 |
| CrossWOZ-IC_test | 4238 | 9 |
| YiwiseIC_FS_train | 1340 | 134 |
| SMP2017-ECDT_FS_train | 618 | 31 |
| CrossWOZ-IC_FS_train | 1200 | 6 |

## 4.2. Training Details

In this paper, the experimental environment is Ubantu18.0, GPU is used for training, and the development language is Python. We use simbert-base-chinese[24] as the baseline model and choose EDA, Mixup-Transformer and TMix as the control group. In this paper, the accuracy on the test set is used as the evaluation result. For not using the recently proposed DMix and DoubleMix, it is because traditional data enhancement operations are incorporated in both studies. Regarding the configuration of EDA, in this paper, the number of generated enhancement samples is set to 8 and the modification ratio alpha=0.1 in this paper; the hyperparameter settings regarding Mixup-Transformer and TMix are adopted from the original paper.

## 4.3. Results

### 4.3.1. Full-sample experiments

Table 2 Results of full-sample comparison experiments

| Models | YiwiseIC | SMP2017-ECDT | CrossWOZ-IC |
|---|---|---|---|
| **Simbert(baseline)** | 91.79 | 95.05 | 95.14 |
| **EDA** | 90.28 | 93.40 | 95.12 |
| **Mixup-Transformer** | 92.50 | 94.45 | 95.38 |
| **TMix** | 92.61 | 95.35 | 95.63 |
| **MSMix-base** | 94.09 | 95.36 | 95.64 |
| **MSMix-A** | 93.53 | **95.80** | **95.87** |
| **MSMix-B** | **94.30** | 95.35 | 95.75 |

In order to verify the effectiveness of the MSMix method proposed in this paper for text classification tasks, four models, MSMix, TMix, Mixup-Transformer, and EDA, are compared in experiments on three datasets. The experimental results are shown in Table 2.

The classification accuracy of MSMix on three Chinese datasets, YiwiseIC, SMP2017-ECDT and CrossWOZ-IC, reached 94.30%, 95.80% and 95.87%, respectively, improving 2.51, 0.75 and 0.73 percentage points, respectively, compared with the baseline model. It indicates that MSMix is effective. the optimal performance of MSMix outperforms TMix on YiwiseIC, SMP2017-ECDT and CrossWOZ-IC datasets. MSMix outperforms EDA on all three datasets. Experiments find that EDA does not play a positive role for some specific Chinese intent recognition. EDA decreases significantly on YiwiseIC and SMP2017-ECDT datasets than on CrossWOZ-IC dataset. After observing the datasets, we found that the text of YiwiseIC and SMP2017-ECDT datasets is shorter, and the random modification operation of EDA may destroy the semantics in a larger way, thus reducing the final effect.

### 4.3.2. Experiments under small sample configuration

The experiments in this paper set up small sample scenarios by (1) for the training set in the YiwiseIC dataset, 10 samples are randomly selected by each class to form the training set of YiwiseIC_FS; (2) for the training set in the SMP2017-ECDT dataset, 20 samples are randomly selected by each class to form the training set of SMP2017-ECDT_FS; (3) The CrossWOZ-IC dataset is extracted from the CrossWOZ dialogue data, and the intention label for each text sentence in CrossWOZ is given in combination with the contextual dialogue information, so the experiments in this paper eliminate the data from three obviously unclean classes when setting up the small sample configuration, and 200 samples of data are taken from each of the remaining 6 classes to form the training set of CrossWOZ-IC_FS.

The classification accuracies of MSMix on three Chinese datasets, YiwiseIC_FS, SMP2017-ECDT_FS and CrossWOZ-IC_FS, reached 82.80%, 91.45% and 92.62%, respectively, improving 1.63, 1.64 and 1.5 percentage points, respectively, compared with the baseline model. It indicates that MSMix is still effective in small sample scenarios. Compared with the experimental results of the full sample, the improvement of the model in this paper is more significant under the small sample configuration on the SMP2017-ECDT and CrossWOZ datasets. Meanwhile, the optimal effect among the three MSMix strategies proposed in this paper is better than the other models. The specific experimental results are shown in Table 3.

Table 3 Comparison of experimental results in small sample configurations

| Models | YiwiseIC_FS | SMP2017-ECDT_FS | CrossWOZ-IC_FS |
|---|---|---|---|
| Simbert(baseline) | 81.17 | 89.81 | 91.12 |
| EDA | 79.73 | 90.85 | 91.77 |
| Mixup-Transformer | 80.04 | 90.85 | 90.65 |
| TMix | 81.40 | 90.10 | 92.00 |
| MSMix-base | 81.98 | 90.40 | 92.00 |
| MSMix-A | **82.80** | 90.85 | **92.62** |
| MSMix-B | 82.54 | **91.45** | 92.55 |

### 4.3.3. Analysis on the replacement strategy in the MSMix

To verify whether the conjecture proposed in 3.2 is feasible, the proposed MSMix-A and MSMix-B are experimented on three Chinese datasets, YiwiseIC, SMP2017-ECDT and CrossWOZ, and in small sample configurations, respectively. As shown in Table 2 and Table 3. MSMix-base, MSMix-A, and MSMix-B outperform other methods in different configurations on the three datasets. However, for the three strategies of selecting replacement dimensions, no single strategy consistently achieves the best performance across all experiments. The following is a brief analysis:

According to the experimental results, the MSMix method performs better than the baseline text Mixup method, i.e., TMix. The reason for this is that the virtual samples generated by the Mixup method are linear values between two real samples in the hidden representation space, which are continuous. Although it greatly enriches the hidden representation space, the natural discreteness of text makes the method proposed in this paper better suited to this characteristic. The MSMix method combines the hidden representation as a finite number of dimensions. Although the specific meaning of each dimension is currently unclear, it preserves the integrity of local text meaning through finite dimensions, thus better combining reasonable samples.

The strategies for selecting replacement dimensions in MSMix-A and MSMix-B are based on an intuitive assumption that the larger the absolute value of a dimension, the more important it is for the current text semantic representation. The random replacement strategy of MSMix-base may replace important features of the sample with unimportant features, which is obviously not an ideal synthetic data. It is similar to the input space sample being composed of some meaningless particles or adverbs to form a sentence. Therefore, we proposes MSMix-A and MSMix-B.

The MSMix-A method replaces the dimension with a larger absolute value obtained by multiplying the corresponding dimensions of two hidden representations, thereby avoiding the aforementioned problem to some extent. However, the following situations may still occur among these larger values: (1) both multiplication values are relatively large, causing an important feature in $h_i^k$ to be replaced with a more important feature in $h_j^k$; (2) values in $h_i^k$ are large while those in $h_j^k$ are small, meaning an important feature in $h_i^k$ is replaced with an unimportant feature in $h_j^k$; (3) values in $h_i^k$ are small while those in $h_j^k$ are large, causing an unimportant feature in $h_i^k$ to be replaced with an important feature in $h_j^k$. Because of the existence of these three situations, the replacement strategy also has a certain degree of randomness. As for MSMix-B, the intention of this strategy was to replace the dimensions of unimportant features in A with the corresponding dimensions in B that are relatively important. This method may result in situations where the importance of the replaced feature becomes stronger than the original important feature, therefore also having a certain degree of instability.

Although MSMix-base, MSMix-A, and MSMix-B exhibit unstable performance when used individually, using these three strategies in combination can achieve better performance than other data augmentation methods.

### 4.3.4. Selecting hidden layers

Research[25] has shown that different hidden layers in the Bert model have different representation capabilities for different features. For example, the 9th layer of Bert is more focused on semantic representation, while the 3rd layer is more focused on the length of the sentence. Some previous work[4, 17, 18] on Mixup has also studied the selection of which hidden layer output in the Bert model to use as the object of Mixup, based on the findings of literature[25].

From Table 2 and Table 3, we can find that TMix (randomly selecting layers in a subset of hidden layers for Mixup) will be better than Mixup-Transformer (fixedly selecting the last hidden layer for Mixup operation). Thus, in this paper, we experimentally set up two sets of hidden layers {k=12} and {k<12}. In this paper, the output of the last hidden layer of Simbert is Mixup as k=12, and the middle hidden layer of Simbert is randomly selected for Mixup as random k. The results of the experiments are shown in Tables 4 and 5. It is found that most cases should randomly select the middle hidden layer instead of fixedly selecting the last hidden layer for Mixup operation, which will bring better results.

Table 4 Experimental results of different hidden layers for full samples

| Models | YiwiseIC | SMP2017-ECDT | CrossWOZ-IC |
|---|---|---|---|
| MSMix-base(k=12) | 91.98 | 95.65 | 95.42 |
| MSMix-base(random k) | 94.09 | 95.35 | 95.64 |
| MSMix-A(k=12) | 91.76 | 95.20 | 95.30 |
| MSMix-A(random k) | 93.53 | 95.80 | 95.87 |
| MSMix-B(k=12) | 92.75 | 95.35 | 95.39 |
| MSMix-B(random k) | 94.30 | 95.35 | 95.75 |

Table 5 Experimental results of different hidden layers for small samples

| Models | YiwiseIC_FS | SMP2017-ECDT_FS | CrossWOZ-IC_FS |
|---|---|---|---|
| **MSMix-base(k=12)** | 81.72 | 90.70 | 91.48 |
| **MSMix-base(random k)** | 81.98 | 90.40 | 92.00 |
| **MSMix-A(k=12)** | 78.94 | 90.55 | 91.96 |
| **MSMix-A(random k)** | 82.80 | 90.85 | 92.62 |
| **MSMix-B(k=12)** | 80.49 | 90.55 | 95.35 |
| **MSMix-B(random k)** | 82.54 | 91.45 | 95.75 |

# 5. CONCLUSION

We propose a simple and effective interpolation-based data enhancement method: MSMix, which performs dimensional replacement of the hidden representation of a layer after two samples are input to the deep neural network to obtain the new hidden representation passed to the subsequent layers for computation, and propose three different strategies for dimensional replacement. The experimental results on three Chinese intention recognition datasets show that the proposed method can improve the robustness of the deep neural network model and help reduce the problem of overfitting in small sample scenarios, and the proposed MSMix outperforms other existing data enhancement methods on three Chinese datasets and achieves the optimal performance. More effective and stable dimensional replacement strategies will be explored later.

## ACKNOWLEDGEMENTS

This work is supported by Hangzhou Yiwise Intelligence Technology Co., Ltd. Project of Hangzhou Leading Innovative Entrepreneurial Team(201920110028) and Hangzhou Major Scientific and Technological Innovation Project on Artificial Intelligence(2022AIZD0093).

## REFERENCES

[1] J. Devlin, M.-W. Chang, K. Lee and K. Toutanova, "BERT: Pre-training of Deep Bidirectional Transformers for Language Understanding," in NAACL-HLT (1), 2019, pp. 4171-4186.
[2] Y. Wang, Q. Yao, J. T. Kwok and L. M. Ni, "Generalizing from a Few Examples: A Survey on Few-shot Learning," ACM Comput. Surv., vol. 53, no. 3, pp. 63:1-63:34, 2021.
[3] J. W. Wei and K. Zou, "EDA: Easy Data Augmentation Techniques for Boosting Performance on Text Classification Tasks," in EMNLP/IJCNLP (1), 2019, pp. 6381-6387.
[4] J. Chen, Z. Yang and D. Yang, "MixText: Linguistically-Informed Interpolation of Hidden Space for Semi-Supervised Text Classification," in ACL, 2020, pp. 2147-2157.
[5] V. Gupta, Z. Li, A. Kortylewski, C. Zhang, Y. Li, and A. L. Yuille, "SwapMix: Diagnosing and Regularizing the Over-Reliance on Visual Context in Visual Question Answering," in IEEE/CVF Conference on Computer Vision and Pattern Recognition (CVPR), 2022, pp. 5068-5078.
[6] V. Verma, A. Lamb, C. Beckham, A. Najafi, I. Mitliagkas, D. Lopez-Paz, and Y. Bengio, "Manifold Mixup: Better Representations by Interpolating Hidden States," in Proc. of the 36th International Conference on Machine Learning (ICML), 2019, pp. 6438-6447.
[7] S. Yang, W. Xiao, M. Zhang, S. Guo, J. Zhao, and F. Shen, "Image Data Augmentation for Deep Learning: A Survey," arXiv preprint arXiv:2204.08610, 2022.
[8] R. Sennrich, B. Haddow, A. Birch, "Improving Neural Machine Translation Models with Monolingual Data," in Proceedings of the 54th Annual Meeting of the Association for Computational Linguistics (Volume 1: Long Papers), 2016, pp. 86-96.
[9] Ng, Nathan, Cho, Kyunghyun, and Ghassemi, Marzyeh. "SSMBA: Self-Supervised Manifold Based Data Augmentation for Improving Out-of-Domain Robustness." EMNLP (1) 2020: 1268-1283.
[10] A. Anaby-Tavor, B. Carmeli, E. Goldbraich, A. Kantor, G. Kour, S. Shlomov, N. Tepper, and N. Zwerdling, "Do Not Have Enough Data? Deep Learning to the Rescue!," in Proceedings of the AAAI Conference on Artificial Intelligence, 2020, pp. 7383-7390.
[11] B. Guo, S. Han, and H. Huang, "Selective Text Augmentation with Word Roles for Low-Resource Text Classification," CoRR, vol. abs/2209.01560, 2022. [Online]. Available: https://arxiv.org/abs/2209.01560.

[12] Xie, Q., Dai, Z., Hovy, E. H., Luong, M. T., & Le, Q. V. (2019). Unsupervised Data Augmentation. arXiv preprint arXiv:1904.12848.
[13] H. Zhang, M. Cissé, Y.N. Dauphin and D. Lopez-Paz, "mixup: Beyond Empirical Risk Minimization," in ICLR (Poster), 2018.
[14] S. Yun, D. Han, S. Chun, S. J. Oh, Y. Yoo, and J. Choe, "CutMix: Regularization Strategy to Train Strong Classifiers With Localizable Features," in Proceedings of the IEEE/CVF International Conference on Computer Vision (ICCV), 2019, pp. 6022-6031.
[15] S. Yoon, G. Kim, and K. Park, "SSMix: Saliency-Based Span Mixup for Text Classification," in Proceedings of the 2021 Conference of the Association for Computational Linguistics and the 9th International Joint Conference on Natural Language Processing (ACL-IJCNLP), Findings, 2021, pp. 3225-3234.
[16] L. Sun, C. Xia, W. Yin, T. Liang, P. S. Yu, and L. He, "Mixup-Transformer: Dynamic Data Augmentation for NLP Tasks," in Proceedings of the 28th International Conference on Computational Linguistics (COLING), 2020, pp. 3436-3440.
[17] R. Sawhney, M. Thakkar, S. Pandit, R. Soun, D. Jin, D. Yang, and L. Flek, "DMix: Adaptive Distance-aware Interpolative Mixup," in Proceedings of the 60th Annual Meeting of the Association for Computational Linguistics (ACL), vol. 2, 2022, pp. 606-612.
[18] H. Chen, W. Han, D. Yang, and S. Poria, "DoubleMix: Simple Interpolation-Based Data Augmentation for Text Classification," in Proceedings of the 29th International Conference on Computational Linguistics (COLING), 2022, pp. 4622-4632.
[19] L. Carratino, M. Cissé, R. Jenatton, and J.-P. Vert, "On Mixup Regularization," CoRR, vol. abs/2006.06049, 2020. [Online]. Available: https://arxiv.org/abs/2006.06049.
[20] G. E. Hinton, N. Srivastava, A. Krizhevsky, I. Sutskever, and R. Salakhutdinov, "Improving neural networks by preventing co-adaptation of feature detectors," CoRR, vol. abs/1207.0580, 2012. [Online]. Available: https://arxiv.org/abs/1207.0580.
[21] S. Hou and Z. Wang, "Weighted Channel Dropout for Regularization of Deep Convolutional Neural Network," in Proceedings of the Thirty-Third AAAI Conference on Artificial Intelligence (AAAI), 2019, pp. 8425-8432.
[22] W. Zhang, Z. Chen, W. Che, G. Hu, and T. Liu, "The First Evaluation of Chinese Human-Computer Dialogue Technology," CoRR, vol. abs/1709.10217, 2017. [Online]. Available: https://arxiv.org/abs/1709.10217.
[23] Q. Zhu, K. Huang, Z. Zhang, X. Zhu, and M. Huang, "CrossWOZ: A Large-Scale Chinese Cross-Domain Task-Oriented Dialogue Dataset," Transactions of the Association for Computational Linguistics (TACL), vol. 8, pp. 281-295, 2020.
[24] J. Su, "SimBERT: Integrating Retrieval and Generation into BERT," [Online]. Available: https://github.com/ZhuiyiTechnology/simbert, Jan. 26, 2021.
[25] G. Jawahar, B. Sagot, and D. Seddah, "What Does BERT Learn about the Structure of Language?" in Proceedings of the 57th Annual Meeting of the Association for Computational Linguistics (ACL), vol. 1, 2019, pp. 3651-3657.